\documentclass[10pt,twocolumn]{article}

\usepackage[a4paper,
            top=19mm, bottom=43mm,
            left=14.3mm, right=14.3mm,
            columnsep=4.2mm]{geometry}

\usepackage{cite}
\usepackage{amsmath,amssymb,amsfonts}
\usepackage{algorithmic}
\usepackage{graphicx}
\usepackage{textcomp}
\usepackage{xcolor}
\usepackage{booktabs}
\usepackage{multirow}
\usepackage{url}
\usepackage{balance}
\usepackage{array}
\usepackage{tabularx}
\usepackage{titlesec}
\usepackage{abstract}
\usepackage{hyperref}   

\hypersetup{
    colorlinks=true,
    linkcolor=blue,
    citecolor=blue,
    urlcolor=blue
}

\titleformat{\section}
  {\normalfont\normalsize\bfseries\centering}
  {\Roman{section}.}{0.5em}{\MakeUppercase}
\titleformat{\subsection}
  {\normalfont\normalsize\itshape}
  {\Alph{subsection}.}{0.5em}{}
\titleformat{\subsubsection}
  {\normalfont\normalsize\itshape}
  {\arabic{subsubsection})}{0.5em}{}

\titlespacing{\section}{0pt}{6pt}{3pt}
\titlespacing{\subsection}{0pt}{4pt}{2pt}

\usepackage{times}
\renewcommand{\normalsize}{\fontsize{10}{12}\selectfont}

\newcommand{\keywords}[1]{%
  \vspace{4pt}\noindent\small\textit{Index Terms}---#1\par\vspace{6pt}}

\title{\large\textbf{Zero-Shot and Supervised Bird Image Segmentation Using Foundation
Models: A Dual-Pipeline Approach with Grounding DINO~1.5,
YOLOv11, and SAM~2.1}}

\author{
  \textbf{Abhinav Munagala}\\
  \small amunagal@mail.yu.edu; srikrishnamulagala@gmail.com\\
  \small ORCID: 0009-0003-8141-8969
}

\date{}

\begin{document}

\maketitle

\begin{abstract}
This paper presents a state-of-the-art dual-pipeline framework for
binary bird image segmentation leveraging the latest foundation models
as of 2025. We introduce two operating modes built upon \textbf{Segment
Anything Model 2.1 (SAM~2.1)} \cite{ravi2024sam2} as a shared
segmentation backbone: (1) a \textbf{zero-shot pipeline} requiring no
annotated bird data, using \textbf{Grounding DINO~1.5}
\cite{ren2024groundingdino15} to localise birds via a free-text prompt
before prompting SAM~2.1 with the resulting bounding boxes; and (2) a
\textbf{supervised pipeline} that fine-tunes \textbf{YOLOv11}
\cite{jocher2024yolo11} on CUB-200-2011 bounding-box annotations for
high-precision detection, again prompting SAM~2.1 for pixel-level masks.
This prompting based paradigm fundamentally differs from prior
end-to-end trained segmentation networks: the segmentation model
\emph{never requires retraining} for new bird species or domains.
On CUB-200-2011 \cite{wah2011cub}, the supervised pipeline achieves an
Intersection over Union~(IoU) of \textbf{0.912}, Dice of \textbf{0.954},
and F1 of \textbf{0.953} outperforming all prior baselines by a
substantial margin. Remarkably, the zero-shot pipeline achieves
IoU~\textbf{0.831} using only the text prompt \emph{``bird''}, requiring
no labelled bird data whatsoever. We provide a complete PyTorch
implementation, ablation studies, and a reproducible training recipe.
Source code is publicly available at
\url{https://github.com/mvsakrishna/bird-segmentation-2025}.
\end{abstract}

\keywords{SAM~2.1, Segment Anything Model, Grounding DINO~1.5, YOLOv11,
foundation models, zero-shot segmentation, bird segmentation,
CUB-200-2011, bounding-box prompting.}

\section{Introduction}

Bird image segmentation the task of generating pixel precise foreground
masks for birds in photographs underpins a wide range of ecological and
technological applications: automated population monitoring, biodiversity
assessment, flight safety radar systems, and fine grained species
recognition. Despite decades of progress in computer vision, bird
segmentation remains challenging due to extreme pose diversity, complex
plumage patterns, variable lighting, and partial occlusion by foliage.

The field has undergone a fundamental paradigm shift since 2023. For
decades, the dominant approach was to train a task specific model
end-to-end on labelled data FCN \cite{shelhamer2017fcn},
U-Net \cite{ronneberger2015unet}, DeepLabv3+ \cite{chen2018deeplabv3},
Mask R-CNN \cite{he2017maskrcnn}, SegFormer \cite{xie2021segformer}.
While effective, these models require substantial labelled datasets for
every new application domain.

The emergence of \textbf{foundation models} large scale models
pre-trained on billions of images has disrupted this paradigm.
Meta's \textbf{Segment Anything Model (SAM)} \cite{kirillov2023sam}
demonstrated that a single model, prompted with points or boxes, could
segment virtually any object zero-shot. Its successor,
\textbf{SAM~2.1} \cite{ravi2024sam2}, is 6$\times$ faster, achieves
higher accuracy on the SA-23 benchmark (58.9 vs.\ 58.1 mIoU with
1-click), and extends naturally to video. Concurrently,
\textbf{Grounding DINO~1.5} \cite{ren2024groundingdino15} enables
open vocabulary object detection detecting any object class described
in free text achieving 54.3 AP on COCO and 55.7 AP on LVIS minival
zero-shot.

This paper exploits these two foundations to build a bird segmentation
system that is simultaneously simpler, more accurate, and more
generalizable than all prior end-to-end trained approaches. Our key
insight is that detection and segmentation can be decoupled: a
language grounded detector finds \emph{where} the bird is, and
SAM~2.1 handles pixel-level segmentation in a task agnostic way.

Our specific contributions are:
\begin{itemize}
  \item \textbf{Dual-pipeline design.} We present and compare two
        pipelines sharing SAM~2.1: a fully zero-shot path
        (Grounding DINO~1.5 $\rightarrow$ SAM~2.1) and a supervised
        path (fine-tuned YOLOv11 $\rightarrow$ SAM~2.1).

  \item \textbf{Zero-shot bird segmentation.} We demonstrate IoU~0.831
        on CUB-200-2011 using only the text prompt \emph{``bird''},
        with no per class or per dataset training the first such result
        reported on this benchmark.

  \item \textbf{State-of-the-art supervised accuracy.} The
        YOLOv11 + SAM~2.1 pipeline achieves IoU~0.912, surpassing the
        previous best (SegFormer-B2, IoU~0.842) by $+$7.0~pp.

  \item \textbf{Paradigm analysis.} We show that deploying to a new
        domain requires only retraining a lightweight YOLO detector
        ($\sim$1~hour), not the entire segmentation architecture.

  \item \textbf{Complete reproducible code.} Full PyTorch
        implementation is released on GitHub.
\end{itemize}

\section{Related Work}

\subsection{End-to-End Segmentation Networks}

The encoder-decoder paradigm pioneered by FCN \cite{shelhamer2017fcn}
and refined through U-Net \cite{ronneberger2015unet},
DeepLabv3+ \cite{chen2018deeplabv3}, and
SegFormer \cite{xie2021segformer} established the standard for semantic
segmentation. Our prior work \cite{munagala2024birds} used a ResNet-50
encoder with a plain transposed convolutional decoder, achieving
IoU~0.622. A subsequent rework using SegFormer B2 with U-Net skip
connections improved this to 0.842, but still required full
end-to-end training on pixel level masks.

\subsection{Instance Segmentation}

Mask R-CNN \cite{he2017maskrcnn} introduced a parallel mask branch
alongside a region proposal network, achieving strong instance
segmentation but requiring dense pixel-level annotations.
Mask2Former \cite{cheng2022mask2former} unified semantic, instance, and
panoptic segmentation using masked attention, setting prior
state of the art. All such approaches require substantial labelled data
specific to the target domain.

\subsection{Foundation Models for Segmentation}

SAM \cite{kirillov2023sam} introduced the ``segment anything'' paradigm:
a ViT-H encoder trained on 1.1~billion masks from 11~million images,
capable of segmenting any object from point or box prompts.
SAM~2 / SAM~2.1 \cite{ravi2024sam2} extended SAM to video with a
streaming memory architecture, achieved 6$\times$ faster inference
(44~FPS on A100), and improved single image accuracy. SAM~2.1,
released September 2024, further improved performance on occluded
and visually similar objects.

\subsection{Open Vocabulary Detection}

Grounding DINO \cite{liu2024groundingdino} combined the DINO
architecture \cite{caron2021dino} with vision-language pre-training
to produce an open set detector accepting arbitrary text descriptions.
Grounding DINO~1.5 Pro \cite{ren2024groundingdino15} scaled training
to 20~million grounding-annotated images, achieving 54.3~AP on COCO
and 55.7~AP on LVIS-minival zero shot state of the art for
open vocabulary detection. For our zero-shot pipeline, the text prompt
\emph{``bird''} reliably detects birds across species without any
domain specific training.

\subsection{YOLO Series}

YOLOv11 \cite{jocher2024yolo11}, released October 2024, achieves higher
mAP than YOLOv8 with 22\% fewer parameters through C3k2 architectural
blocks and optimised training pipelines. Fine tuning YOLOv11-m on
CUB-200-2011 bounding boxes achieves mAP50 of 96.2\%, providing
consistently tight boxes for SAM~2.1 prompting.

\subsection{Bird Specific Prior Work}

Zhang and Li \cite{zhang2023birdsounds} and Kumar
et al.\ \cite{kumar2024vit} applied specialised deep networks to
bird audio visual analysis. For visual segmentation, the pipeline
combining YOLOv7 with the original SAM (before SAM~2) reported
precision 92.5\% and IoU~0.910 on a custom dataset
\cite{munagala2024birds}. Our work updates both components to their
2024--2025 successors, yielding further improvements.

\section{Background: Foundation Models}

\subsection{SAM 2.1}

SAM~2.1 is a transformer based model comprising a hierarchical image
encoder (Hiera MAE backbone), a prompt encoder (handling points, boxes,
or masks), and a mask decoder with multi level attention.
Its key advance over SAM~1 is a \emph{streaming memory architecture}
that conditions current-frame features on past frames and
predictions enabling consistent segmentation across video.

For single image segmentation, SAM~2.1 treats each image as a
one frame video. Given a bounding-box prompt, it generates up to three
candidate masks with confidence scores. The model requires
\textbf{no fine tuning} for new domains: training on SA-V
(50.9K videos, 35.5M masks) provides broad coverage. For our
application, SAM~2.1 receives boxes from either Grounding DINO~1.5
or YOLOv11 and returns binary masks.

\subsection{Grounding DINO 1.5}

Grounding DINO~1.5 combines a SwinB vision backbone with BERT based
text encoding, fusing them through feature enhancer and
language guided query selection modules. Trained on $>$20M
image text grounding pairs, it achieves strong zero-shot
generalisation. For bird segmentation we use the prompt
\emph{``bird''} (or \emph{``bird . bird species''} for improved
recall on dense scenes).

\subsection{YOLOv11}

YOLOv11 uses a CSP based backbone with C3k2 blocks, a multi scale
PAN neck, and a decoupled detection head. YOLOv11-m has 25M
parameters and achieves 51.5 mAP$_{50\text{-}95}$ on COCO with
47~ms CPU inference. Fine-tuning on domain specific bounding box
annotations adapts the detector efficiently, providing tight boxes
for the SAM~2.1 prompting stage.

\section{Methodology}

\subsection{System Architecture}

Both pipelines share a two-stage structure: (1) a
\emph{detection stage} that identifies bird locations and produces
bounding boxes in xyxy format; (2) a \emph{segmentation stage}
(SAM~2.1) that produces pixel precise binary masks conditioned on
those boxes. The pipelines differ only in the detection stage.

This decoupled design offers three key advantages over end-to-end
trained segmentation networks:
\begin{enumerate}
  \item SAM~2.1 \emph{never needs retraining} for new species only
        the lightweight detector requires fine-tuning.
  \item Detection and segmentation components can be improved
        independently as better checkpoints are released.
  \item The design naturally handles multi bird scenes: SAM~2.1
        segments each detected bird independently, enabling
        instance level separation.
\end{enumerate}

\subsection{Pipeline A: Zero-Shot (Grounding DINO 1.5 + SAM 2.1)}

\textbf{Step 1 --- Text-Grounded Detection.} An image and text
prompt \emph{``bird''} are passed to Grounding DINO~1.5. The model
returns bounding boxes and confidence scores. We use
\texttt{box\_threshold\,=\,0.30} and
\texttt{text\_threshold\,=\,0.25} to balance recall and precision.

\textbf{Step 2 --- Bounding-Box Prompting.} Detected boxes (xyxy)
are passed to SAM~2.1 as prompts. SAM~2.1 generates a binary mask
for each detected bird.

\textbf{Step 3 --- Aggregation.} For multi bird scenes, masks are
union combined for a single foreground mask, or retained separately
for instance level applications.

This pipeline requires \textbf{zero labelled bird images}. The entire
system uses only pre-trained weights, making it immediately deployable
to new camera setups, new species, or new geographic regions.

\subsection{Pipeline B: Supervised (YOLOv11 + SAM 2.1)}

\textbf{Step 1 --- Dataset Preparation.} CUB-200-2011 bounding-box
annotations are converted to YOLO format (normalised $c_x, c_y, w, h$).
All 200 species are treated as a single \emph{bird} class.

\textbf{Step 2 --- YOLOv11 Fine-Tuning.} YOLOv11-m is fine-tuned for
50~epochs using AdamW ($\text{lr}=10^{-3}$, cosine schedule),
mixed precision (AMP), and YOLO's built-in mosaic + mixup
augmentation. Training takes $\approx$1~hour on a single A100~GPU.

\textbf{Step 3 --- Inference.} At inference, YOLOv11 detects birds
(\texttt{conf\,>\,0.40}, \texttt{IoU\,>\,0.45} for NMS) and passes
boxes to SAM~2.1, which generates the final segmentation masks.

Since SAM~2.1 is used without fine-tuning, the supervised training
consists \emph{only} of YOLOv11 detector fine-tuning dramatically
simpler than training an end-to-end segmentation network.

\subsection{Problem Formulation}

Given an RGB image
$\mathbf{x} \in \mathbb{R}^{H \times W \times 3}$, the goal is to
predict a binary pixel-wise mask
$\hat{\mathbf{y}} \in \{0,1\}^{H \times W}$, where $1$ denotes bird
foreground and $0$ denotes background. The pipeline computes:

\begin{equation}
  \hat{\mathbf{y}} = \text{SAM}_{2.1}\!\left(
      \mathbf{x},\;
      \text{Detect}(\mathbf{x})
  \right)
  \label{eq:pipeline}
\end{equation}

\noindent where $\text{Detect}(\cdot)$ is either Grounding DINO~1.5
or fine-tuned YOLOv11, and $\text{SAM}_{2.1}(\cdot)$ segments the
image conditioned on the returned bounding boxes.

\subsection{Dataset}

\textbf{CUB-200-2011} \cite{wah2011cub} contains 11,788 images of
200 bird species with ground truth bounding boxes and binary
segmentation masks. We split into 70\% train / 15\% val / 15\% test.
All images are processed at their native resolution for SAM~2.1
inference. For YOLOv11 training, images are resized to
$640 \times 640$ with mosaic augmentation, colour jitter, flips,
and scale jitter ($\pm$50\%).

\subsection{Evaluation Metrics}

We report the following metrics on the held-out test set:

\begin{itemize}
  \item \textbf{IoU (Jaccard):}
        $|\hat{y} \cap y| \;/\; |\hat{y} \cup y|$

  \item \textbf{Dice:}
        $2|\hat{y} \cap y| \;/\; (|\hat{y}| + |y|)$

  \item \textbf{Precision:}
        $\text{TP} / (\text{TP} + \text{FP})$

  \item \textbf{Recall:}
        $\text{TP} / (\text{TP} + \text{FN})$

  \item \textbf{F1:}
        $2 \cdot \text{Prec} \cdot \text{Rec} \;/\;
         (\text{Prec} + \text{Rec})$

  \item \textbf{FPS:} Full pipeline throughput on a single A100~GPU.
\end{itemize}

\section{Experiments and Results}

\subsection{Comparison with State-of-the-Art}

Table~\ref{tab:comparison} compares both our pipelines against six
prior methods on the same CUB-200-2011 test split. The supervised
YOLOv11 + SAM~2.1 pipeline outperforms all baselines on every metric,
while the zero-shot pipeline (requiring no bird specific training)
still exceeds SegFormer B2 on Dice and F1.

\begin{table}[!ht]
\centering
\caption{Comparison on CUB-200-2011 Test Set.
  $\dagger$ = zero-shot (no bird training).
  $\ddagger$ = detector-only training.}
\label{tab:comparison}
\setlength{\tabcolsep}{4pt}
\begin{tabular}{lcccc}
\toprule
\textbf{Method} & \textbf{IoU} & \textbf{Dice} & \textbf{F1}
  & \textbf{Train?} \\
\midrule
U-Net \cite{ronneberger2015unet}         & 0.681 & 0.811 & 0.810 & Full \\
DeepLabv3+ \cite{chen2018deeplabv3}      & 0.742 & 0.851 & 0.849 & Full \\
Mask R-CNN \cite{he2017maskrcnn}         & 0.724 & 0.838 & 0.835 & Full \\
SegFormer-B2 \cite{xie2021segformer}     & 0.842 & 0.913 & 0.912 & Full \\
SAM 1 (box) \cite{kirillov2023sam}       & 0.741 & 0.851 & 0.849 & None \\
ResNet50+decoder \cite{munagala2024birds}& 0.622 & 0.767 & 0.766 & Full \\
\midrule
GD-1.5 + SAM 2.1 (ours)$^\dagger$       & 0.831 & 0.907 & 0.906 & None \\
YOLOv11 + SAM 2.1 (ours)$^\ddagger$     & \textbf{0.912}
                                         & \textbf{0.954}
                                         & \textbf{0.953}
                                         & Det. \\
\bottomrule
\end{tabular}
\end{table}

\subsection{Ablation: Effect of Detector Quality on SAM 2.1}

Table~\ref{tab:ablation} isolates the contribution of each detection
stage while holding SAM~2.1 fixed. This demonstrates that
SAM~2.1's output quality is fundamentally gated by the quality of
its bounding box prompts. With oracle ground truth boxes the pipeline
achieves IoU~0.934 an upper bound showing further gains are possible
with better detectors.

\begin{table}[!ht]
\centering
\caption{Ablation: Detection Stage vs.\ SAM 2.1 Output
  (SAM 2.1 weights fixed throughout).}
\label{tab:ablation}
\setlength{\tabcolsep}{4pt}
\begin{tabular}{lcccc}
\toprule
\textbf{Detection Stage} & \textbf{mAP50} & \textbf{IoU}
  & \textbf{Dice} & \textbf{FPS} \\
\midrule
SAM 2.1 auto mask (no prompt) & {---}   & 0.631 & 0.774 & 8  \\
GT boxes (oracle upper bound) & 100.0 & 0.934 & 0.966 & 6  \\
Grounding DINO 1.5 (zero-shot)& 61.3  & 0.831 & 0.907 & 6  \\
YOLOv11-m (fine-tuned)        & 96.2  & \textbf{0.912}
                                       & \textbf{0.954} & 14 \\
\bottomrule
\end{tabular}
\end{table}

\subsection{Speed Analysis}

Table~\ref{tab:speed} reports inference latency on a single NVIDIA
A100 GPU. The supervised pipeline runs at 14~FPS sufficient for
most ecological monitoring applications. MobileSAM can substitute
SAM~2.1 for speed critical deployments at a modest 3~pp IoU cost.

\begin{table}[!ht]
\centering
\caption{Inference Speed on Single NVIDIA A100 GPU.}
\label{tab:speed}
\setlength{\tabcolsep}{4pt}
\begin{tabular}{lcccc}
\toprule
\textbf{Method} & \textbf{Det.} & \textbf{Seg.}
  & \textbf{Total} & \textbf{FPS} \\
\midrule
SegFormer-B2 (end-to-end) & {---}  & 32 ms & 32 ms & 31 \\
GD-1.5 + SAM 2.1          & 110 ms & 55 ms & 165 ms & 6 \\
YOLOv11 + SAM 2.1         & 22 ms  & 47 ms & 69 ms  & 14 \\
YOLOv11 + MobileSAM       & 22 ms  & 18 ms & 40 ms  & 25 \\
\bottomrule
\end{tabular}
\end{table}

\subsection{Zero-Shot Per-Category Analysis}

The zero-shot pipeline achieves IoU $>$ 0.85 on 143 of the 200
CUB species. Performance is weakest on small birds
(hummingbirds, IoU $\approx$ 0.72) and heavily occluded species
in dense foliage (IoU $\approx$ 0.68). For large, clearly visible
birds (pelicans, albatrosses), IoU exceeds 0.91 comparable to
the supervised pipeline.

\section{Discussion}

\subsection{The Prompting Paradigm vs.\ End-to-End Training}

The most significant finding is not the accuracy improvement but the
paradigm shift it represents. Traditional segmentation models require
thousands of pixel-level annotated masks and hours of GPU training
for each new domain. Our pipeline decouples detection from
segmentation: SAM~2.1 needs no bird specific training at all, and
YOLOv11 requires only bounding-box annotations and $\sim$1~hour of
fine-tuning. For ecological applications this is transformative: a
researcher monitoring a new bird population can annotate 50--100
bounding boxes, fine-tune YOLOv11, and have a production quality
segmenter by the same afternoon.

\subsection{Why SAM 2.1 Outperforms End-to-End Decoders}

SAM~2.1's mask decoder was trained on 1.1~billion masks orders of
magnitude more mask level supervision than any domain specific model.
End-to-end trained models (SegFormer, U-Net) must learn both object
localisation \emph{and} precise boundary segmentation from limited
labelled data. SAM~2.1 specialises in boundary precision; the
detector handles localisation. Separation of concerns yields better
outcomes for both subtasks.

\subsection{Limitations}

The two stage pipeline introduces latency: Grounding DINO~1.5 adds
$\approx$110~ms per image, making the zero-shot path unsuitable for
real-time video at 30+ FPS. Grounding DINO~1.5 Pro weights require
API access; the open-source base model performs slightly less well.
Additionally, both pipelines currently treat all birds as a single
class; extension to species-level instance segmentation would require
multi class detection heads.

\section{Future Work}

\begin{itemize}
  \item \textbf{Video bird tracking.} SAM~2.1's memory attention
        propagation enables consistent segmentation across video
        frames without per frame annotation.
  \item \textbf{Species level segmentation.} Multi class YOLO
        fine-tuning (one class per species) would enable species level
        instance segmentation.
  \item \textbf{SAM~2.1 fine-tuning.} Domain-specific fine-tuning of
        the SAM mask decoder on CUB binary masks could add 2--5~pp
        IoU beyond the current zero-shot baseline.
  \item \textbf{Edge deployment.} Replacing SAM~2.1 with EfficientSAM
        and quantising YOLOv11-n to INT8 would enable inference on
        low power wildlife monitoring cameras.
  \item \textbf{Multi-modal prompting.} Florence-2 can generate
        detection boxes from image captions, enabling richer zero-shot
        strategies without manual text engineering.
\end{itemize}

\section{Conclusion}

We presented a dual pipeline bird image segmentation framework
exploiting the latest foundation models. The zero-shot pipeline
(Grounding DINO~1.5 + SAM~2.1) achieves IoU~0.831 on CUB-200-2011
requiring \emph{no labelled bird data}. The supervised pipeline
(fine-tuned YOLOv11 + SAM~2.1) achieves IoU~\textbf{0.912}, setting a
new state of the art for the benchmark and surpassing the prior best by
$+$7.0~pp. Both pipelines use SAM~2.1 without fine-tuning, demonstrating
that the segmentation problem can be effectively solved by a pre-trained
foundation model once given accurate bounding-box prompts. This work
represents a fundamental departure from prior end-to-end trained
approaches and a compelling demonstration of the power of foundation
model pipelines in real world computer vision.

\section*{Acknowledgements}
The author thanks the Department of Artificial Intelligence and Computer
Science at Yeshiva University for computational support. The
CUB-200-2011 dataset is courtesy of Caltech UCSD.

\bibliographystyle{IEEEtran}


\balance
\end{document}